\documentclass[runningheads]{llncs}

\usepackage{eccv}

\usepackage{eccvabbrv}

\usepackage{graphicx}
\usepackage{booktabs}

\usepackage{amsmath,amssymb}
\usepackage{lipsum}
\usepackage{multirow}
\usepackage{multicol}
\usepackage{makecell}
\usepackage{comment}
\usepackage{dcolumn}
\usepackage[normalem]{ulem}
\usepackage{siunitx}
\usepackage[accsupp]{axessibility}  % Improves PDF readability for those with disabilities.

\usepackage{hyperref}

\usepackage{orcidlink}

\newcommand\norm[1]{\left\lVert#1\right\rVert}
\newcolumntype{d}[1]{D..{#1}}
\definecolor{MainPurple}{HTML}{9881FF}
\definecolor{MainPink}{HTML}{FF439F}
\definecolor{MainYellow}{HTML}{FFD500}
\definecolor{MainBlue}{HTML}{87AEFF}
\definecolor{MainOrange}{HTML}{FF8400}

\definecolor{nicegreen}{rgb}{0.1, 0.6, 0.2}

\begin{document}

\title{Spherical World-Locking for Audio-Visual Localization in Egocentric Videos}

\titlerunning{Spherical World-Locking}

\author{Heeseung Yun$^{1,2}$\thanks{Work done during an internship at Meta.} \and Ruohan Gao$^2$ \and Ishwarya Ananthabhotla$^2$ \and \\ Anurag Kumar$^2$ \and Jacob Donley$^2$ \and Chao Li$^2$ \and Gunhee Kim$^1$ \and \\ Vamsi Krishna Ithapu$^2$ \and Calvin Murdock$^2$}
% \inst{1}\orcidlink{0000-1111-2222-3333} \and
% Second Author\inst{2,3}\orcidlink{1111-2222-3333-4444} \and
% Third Author\inst{3}\orcidlink{2222--3333-4444-5555}}

\authorrunning{H. Yun et al.}

\institute{$^1$Seoul National University, $^2$Reality Labs Research at Meta \\ \url{https://hs-yn.github.io/SWL/}}
% \and
% Springer Heidelberg, Tiergartenstr.~17, 69121 Heidelberg, Germany

\maketitle

\begin{abstract}
Egocentric videos provide comprehensive contexts for user and scene understanding, spanning multisensory perception to behavioral interaction.
We propose Spherical World-Locking (SWL) as a general framework for egocentric scene representation, which implicitly transforms multisensory streams with respect to measurements of head orientation.
% within sphere around the wearer.
Compared to conventional head-locked egocentric representations with a 2D planar field-of-view, SWL effectively offsets challenges posed by self-motion, 
% such as drifts and visibility, 
allowing for 
% an intuitive visualization with localized signals and 
improved spatial synchronization 
% among multisensory inputs.
between input modalities.
Using a set of multisensory embeddings on a world-locked sphere, we design a unified encoder-decoder transformer architecture that preserves the spherical structure of the scene representation, without requiring expensive projections between image and world coordinate systems. % image-to-sphere projections.
We evaluate the effectiveness of the proposed framework on multiple benchmark tasks for egocentric video understanding, including audio-visual active speaker localization, auditory spherical source localization, and behavior anticipation in everyday activities. 
 % Multisensory Representation Learning?
\end{abstract}

\section{Introduction}
\label{sec:introduction}

\begin{figure}[t]
    \centering
\includegraphics[trim=0.0cm 0.0cm 0cm 0.0cm,clip,width=1\textwidth]{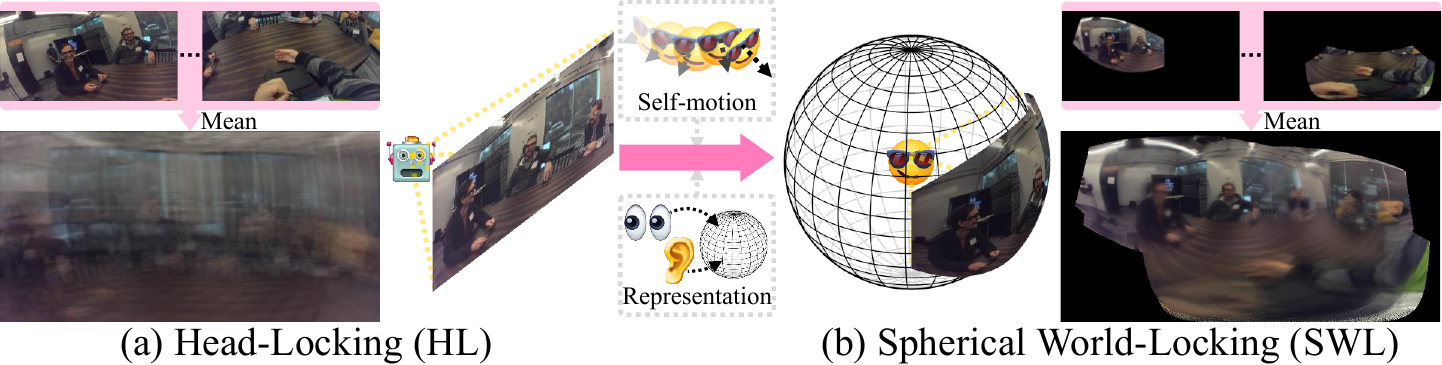}
%\vspace{-15pt}
    \caption{
    The key idea of our framework.
    (a) In conventional \textit{Head-Locked} (HL) frameworks, multisensory observations captured from head-mounted devices are used as-is, where self-motion introduces variability in otherwise static scenes.
    (b) Our Spherical World-Locking (SWL) framework compensates for self-motion with negligible overhead, leading to lower variability and better learnable scene representation. 
    }
    \label{fig:keyidea}
    %\vspace{-10pt}
\end{figure}

%\RGnote{I was told that vspace are not allowed for camera ready (explicitly specified in camera ready instructions), and you might have to remove all and further condense or move something to Supp.}
Egocentric videos provide comprehensive context from an individual's perspective, serving an essential role in user and scene understanding.
Compared to conventional exocentric videos, egocentric videos capture in-the-wild context from a human-centric viewpoint covering daily routine activities and social interactions like conversation.
Therefore, it is paramount to capture the interplay between visual, auditory, and behavioral modalities for comprehensive reasoning tasks in line with how humans would perceive their surroundings~\cite{ohmi1996egocentric,bottini2001cerebral}. % a variety of low-level behaviors like movement or gaze 
% Such a capability is an indispensable component of personalized scene understanding to support applications in augmented and mixed reality. % or social robotics. 
Accordingly, a significant amount of research has been dedicated to exploring the integration of multiple modalities in egocentric videos~\cite{damen2018scaling,kazakos2021little,grauman2022ego4d,lai2023eye,ryan2023egocentric,lv2024aria,tan2024egodistill}.
% ,murdock2024self}.
% In this work, we address localization problems in egocentric videos with audio-visual streams.
In this work, we develop a general framework for multisensory egocentric perception and apply it to audio-visual and behavioral localization problems in egocentric videos.

One of the most distinctive characteristics of egocentric videos is self-motion.
This is best illustrated in a group conversation where people frequently move their heads to engage in various actions such as nodding, making eye contact, or visually exploring their surroundings.
That is, self-motion poses an important challenge for egocentric video understanding, as the relative location of stimuli in a head-locked reference frame would also move accordingly. % with respect to this motion.
Other factors stemming from self-motion, like motion drift and a limited field-of-view, also contribute to the increased complexity. % of videos. %egocentric video understanding.
% Due to its complex nature, 
Therefore, 
self-motion is often treated as a challenge in egocentric video understanding~\cite{liu2022egocentric,jiang2022egocentric,bansal2022my,ryan2023egocentric,mai2023egoloc}.

Nevertheless, self-motion is one of the core components of egocentric scene understanding, acting as a strong proxy of behavior and its underlying intention.
For instance, the internal representations of our perceived surroundings do not change significantly with respect to drastic self-motion and always remain gravity-aligned thanks to behavioral responses  like the vestibulo-ocular~\cite{wallach1940role} and proprioceptive reflexes~\cite{gauthier1995egocentric}.
% Also, humans can
We can also effortlessly coordinate head motion to proactively sharpen our perception of attended contexts~\cite{thurlow1967head, brimijoin2013contribution}.
In other words, humans are efficient stabilizers as well as utilizers of self-motion, 
% self-motion is used stabilize and improve human multisensory perception, 
and we claim that these traits can be beneficially adapted for egocentric video understanding.
% Likewise, it would be beneficial to consolidate such properties into intelligent models for an effective egocentric video understanding by properly leveraging the wearer's motion information.

To this end, 
we introduce \textbf{S}pherical \textbf{W}orld-\textbf{L}ocking (SWL) as a novel framework for integrating self-motion into egocentric videos.
As depicted in Fig.~\ref{fig:keyidea}, 
in contrast to conventional \textit{Head-Locked} frameworks that learn to offset variability from self-motion in raw input streams,
SWL forms a virtual sphere around a person and efficiently transforms audio-visual streams based on measurements of their relative head orientation.
This is generally applicable in egocentric videos by leveraging 
% pose information of the wearer from 
sensors like inertial measurement units 
%that are commonly available
in commercial head-mounted  devices~\cite{damen2018scaling,donley2021easycom,grauman2022ego4d,lv2024aria}.
SWL inherently offsets challenges posed by self-motion such as drift and visibility, while maintaining the strengths of a head-locked representation like compatibility with conventional 
%audio-visual
frameworks.
% It can be formulated either explicitly for an intuitive visualization on a spherical panorama, or implicitly for self-motion-aware multisensory representation learning.
%This property allows for intuitive visualization of egocentric videos by explicitly transforming the input streams on a spherical panorama.

We propose the \textbf{Mu}ltisensory \textbf{S}pherical World-Locked \textbf{T}ransformer (MuST), a novel architecture that incorporates self-motion-aware multisensory inputs for representation learning. % multisensory input streams
% while leveraging self-motion as a useful cue for learning.
% with self-motion compensation while avoiding expensive explicit projection to the sphere.
MuST leverages self-motion 
as a useful cue for learning 
with negligible computational overhead
% in the form of 
via
% position embeddings
spherical position embeddings. % al encodings.
% MuST can directly reflect self-motion during training in the form of position embeddings with negligible overheads.
% We further facilitate the interplay of multisensory inputs on the world-locked sphere by introducing 
To further facilitate the interaction of multisensory inputs on the world-locked sphere,
we introduce
modality-wise self-attention and quaternion-based spatial similarity. 
In addition, multiple classification tokens improve localized predictions and allow for flexible decoding on multiple target domains, \eg, field-of-view, spherical, and pointwise prediction.

We validate the effectiveness of our framework with three representative benchmarks for multisensory egocentric perception and understanding.
First, we outperform prior arts by a significant margin in the audio-visual active speaker localization benchmark on the EasyCom dataset~\cite{donley2021easycom}.
Second, we obtain superior auditory spherical source localization performance on the 
%large-scale 
RLR-CHAT dataset~\cite{murdock2024self,yin2024hearing}.
% of free-form conversations.
Finally, to demonstrate the generalizability of our framework, we extend it to egocentric behavior anticipation in everyday activities on the Aria Everyday Activities (AEA) dataset~\cite{lv2024aria}, where we establish a new benchmark of jointly predicting a set of cohesive behaviors from multisensory contexts.

\begin{figure}[t]
    \centering
    \includegraphics[trim=0.0cm 0.0cm 0cm 0.0cm,clip,width=1\textwidth]{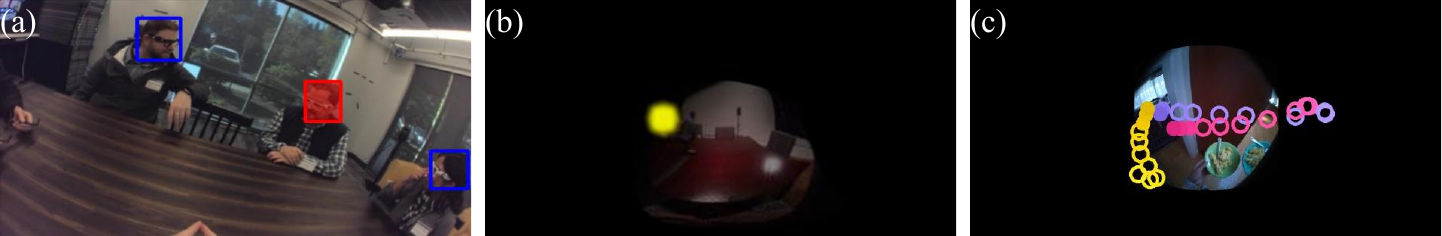}
    % \vspace{-0.2in}
    \caption{Three multisensory localization tasks in egocentric videos that we tackle in this work: (a) audio-visual active speaker localization (\S\ref{subsec:exp_easycom}), (b) auditory spherical source localization (\S\ref{subsec:exp_chat}), and (c) egocentric behavior anticipation (\S\ref{subsec:exp_ariapilot}).}
    \label{fig:task}
    % \vspace{-10pt}
\end{figure}

% \vspace{-0.1in}
\section{Related Works}
\label{sec:relatedworks}
% \vspace{-0.1in}

%\heeseung{Please feel free to include additional paper(s) that you think is missing.}
\textbf{Egocentric Video Understanding}.
A variety of reasoning tasks have been studied from the wearer-centric perspective based on the wearer's spatiotemporal or multimodal context~\cite{damen2018scaling,grauman2022ego4d,grauman2023ego}.
Some prior works focus on improving egocentric human-object interaction by predicting the wearer's hand motion~\cite{liu2020forecasting} or hand segmentation~\cite{jia2022generative} and discerning distracting objects~\cite{wang2021interactive}.
% , amongst others. % active and distracting objects
In addition, information from third-person videos can be transferred to the egocentric domain by means of knowledge distillation~\cite{li2021ego} or temporal alignment without paired data~\cite{xue2023learning}.
Other works learn topological maps with longer temporal dependencies~\cite{nagarajan2020ego} or pre-training with embodied agents~\cite{nagarajan2023egoenv}.
Also, fine-grained temporal relationships among multiple modalities are modeled~\cite{kazakos2019epic,kazakos2021little}.

Egocentric videos with additional modalities measuring the wearer's behavior offer unique challenges in understanding the context. 
% wearer's context.
% The wearer
Pose estimation is one of the primary tasks of egocentric user understanding, making use of dynamic motion signatures~\cite{jiang2017seeing}, body segmentation with motion history~\cite{jiang2021egocentric}, or an intersection between kinematics and dynamics~\cite{luo2021dynamics}.
% The wearer's 
Gaze has been estimated by learning the correlation between global context and local information of visual tokens~\cite{lai2023eye}.
A more recent line of research utilizes the wearer's pose information, \ie, IMU sensors, for efficient action recognition~\cite{tan2024egodistill} or translation to textual description via contrastive pre-training~\cite{moon2023imu2clip}.
% In this work, we propose a novel framework to integrate the wearer's pose with other multimodal observations by compensating for self-motion in egocentric videos.
While previous works make use of \textit{unimodal} models for IMU and train them with cross-modal learning objectives, we directly incorporate the wearer's pose with audio-visual embeddings for \textit{multimodal} models preserving spherical world-locked structure.

% \vspace{0.05in}
\noindent
\textbf{Audio-Visual Localization.}
% Audio-only: directional audio coding with specific microphone array configuration~\cite{pulkki2006directional,ahonen2008directional}, array transfer function~\cite{tourbabin2018space,tourbabin2019direction}
Extensive research has been conducted to exploit audio-visual correspondence for localization in videos in-the-wild with self-supervision~\cite{owens2018audio,arandjelovic2018objects,gao20192,sun2023learning}, cross-modal clustering~\cite{hu2019deep,mo2023audio}, parameter-efficient adaptation~\cite{lin2023vision}, and pixel-level correspondence learning~\cite{zhao2018sound,zhao2019sound,wu2021binaural}, to list a few.
Audio-visual speaker detection is another major task in identifying the coherence between the two modalities under multi-speaker scenarios~\cite{afouras2020self,roth2020ava,kim2021look,tao2021someone,truong2021right}.
Other recent works on audio-visual localization exploit a richer set of modalities to enhance localization capabilities, like question-answer grounding~\cite{yun2021pano}, language-guide separation~\cite{tan2023language}, cross-view consistency of source directions~\cite{chen2023sound}, homography~\cite{huang2023egocentric}, and audio-visual saliency~\cite{xiong2023casp}.

% Huang et al. propose homography-based geometry-aware temporal aggregation to mitigate drifts in egocentric videos.
%videos, 
% Egocentric localization of audio-visual signals poses new challenges to various settings from daily activities to conversations.
Conversations in egocentric videos are often more complex than in conventional videos due to noisy environments and unconstrained multi-speaker interactions.
Jiang et al.~\cite{jiang2022egocentric} combine audio-only and audio-visual networks to perform spherical and inner field-of-view active speaker localization.
Ryan et al.~\cite{ryan2023egocentric} refine this localization capability by only detecting the attended conversation partner, 
% they are listening to, 
while Jia et al.~\cite{jia2024avconv} further propose to predict the complete ego-exo conversational graph from egocentric video.
%propose complete conversation graph prediction.
Closest to our work is~\cite{murdock2024self}, where self-motion behaviors are used as a self-supervised learning objective.
Unlike 
%Spherical World-Locking,
SWL,
the self-supervisory signal cannot solely offset the challenges of self-motion.
% whereas our framework provides a straightforward method to integrate self-motion as inputs directly.

% \vspace{0.05in}
\noindent
\textbf{Spherical Scene Representation.}
There has been a surge of interest in representing spherical data, from modeling 360$^\circ$ videos~\cite{su2016pano2vid,lee2018memory} to the Earth's climate~\cite{mudigonda2017segmenting}.
% Spherical 360$^\circ$
Omnidirectional
videos are generally projected into multiple normal field-of-view images to mitigate distortion and discontinuity~\cite{su2016pano2vid,lee2018memory}.
On the other hand, some prior works propose invariant or equivariant architectures on a sphere~\cite{jaderberg2015spatial,cohen2018spherical,esteves2018learning}, discretization with polyhedral approximation~\cite{eder2020tangent}, or data structures like the spherical binoctree~\cite{jang2022egocentric} and balanced spherical grid~\cite{choi2023balanced} for more faithful representations of the spherical scene.
Other works focus on the transferability of convolutional networks~\cite{su2021learning} or transformers~\cite{yun2022panoramic} from the normal image domain to the 360$^\circ$ domain.
However, egocentric videos are planar and not spherical by nature, making it less practical to adopt their spherical architectures for learning.
Instead, our method interprets egocentric videos on a world-locked sphere while preserving their original format, without requiring additional expensive mechanisms like the Spatial Transformer~\cite{jaderberg2015spatial} to incorporate the spherical nature of egocentric observations.

\section{Spherical World-Locking}
\label{sec:approach_swl}

In conventional head-locked frameworks, egocentric videos are provided as-is, and the model must learn the complex nature of self-motion from end to end.
In contrast, we propose Spherical World-Locking (SWL) to represent videos on a world-locked sphere around the wearer's head, serving as an effective means to model self-motion in egocentric videos.
Since these audio-visual data are not spherical in nature, we first need to establish the connection between the world-locked sphere and audio-visual egocentric streams.
We use multisensory egocentric inputs comprising a video frame $\mathcal{V}$ $(3\times H_v\times W_v)$, the corresponding multichannel audio spectrogram $\mathcal{A}$ $(C_a\times H_a\times W_a)$, and behaviors $\mathcal{B}$ as 3D unit vectors $(N_b\times 3)$ if available.
We consider three different directional behaviors of the wearer, \ie, eye gaze, head orientation, and motion trajectory.

\begin{figure}[t]
    \centering
    % \vspace{-10pt}
\includegraphics[trim=0.0cm 0.0cm 0cm 0.0cm,clip,width=1\textwidth]{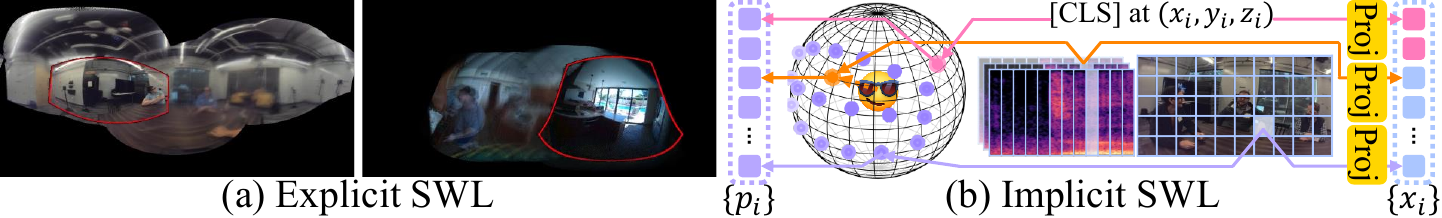}
    % \vspace{-15pt}
    \caption{Comparison of explicit and implicit spherical world-locking.
    While explicit SWL maps the original inputs to the spherical reference frame, implicit SWL retains the original inputs to process position ($\{p_i\}$) and semantic information ($\{x_i\}$) separately.
    }
    \label{fig:swl}
    % \vspace{-10pt}
\end{figure}

As visualized in Fig.~\ref{fig:swl},
% the output of SWL varies with respect to the mapping function we want to obtain, and we discuss two different directions to derive this.
we formulate two different SWL methods that can equivalently represent the world-locked sphere, where each has its distinct advantages.
\textit{Explicit} spherical world-locking (\S\ref{subsec:explicitswl}) maps the original video to a $360^\circ$ panorama, \ie, $f_{EX}: \mathcal{V}\mapsto\mathcal{V}'$, where $\mathcal{V}'$ $(3\times H_p\times W_p)$ is a panoramic image.
Whereas in \textit{implicit} spherical world-locking (\S\ref{subsec:implicitswl}), we keep the original inputs and construct the mapping from patches (for audio-visual inputs) or vectors (for other inputs) to tuples of semantic and position embeddings that encode the corresponding self-motion.

\begin{comment}
\noindent
\begin{minipage}{\textwidth}
\begin{minipage}[b]{\textwidth}
\centering
\begin{tabular}{c|l}
\hline
$\mathcal{A,V,B}$ & Inputs (\textbf{a}udio, \textbf{v}isual, \textbf{b}ehavior) \\
$c_i, a_i, v_i, b_i$ & CLS token, audio, visual and behavior embeddings ($\mathbb{R}^{N_{c,a,v,b}\times d}$) \\
$N, d$  & The number of embeddings, the length of feature dimension \\
$x_i^l$ & The $i$-th semantic embedding at layer $l$ ($\mathbb{R}^{N\times d}=\mathbb{R}^{(N_c+N_a+N_v+N_b)\times d}$) \\
$p_i$ & The position of $i$-th embedding on a world-locked sphere ($\mathbb{R}^{N\times 3}$) \\ \hline
\end{tabular}
\end{minipage}
\end{minipage}
\end{comment}

\subsection{Explicit Spherical World-Locking}
\label{subsec:explicitswl}

We first outline the procedure of placing egocentric videos on a world-locked sphere.
Among the three multisensory inputs, behaviors with direction $\mathcal{B}$ can be trivially placed on a sphere with scalar multiplication.
However, assigning a precise direction to audio inputs $\mathcal{A}$ is difficult. In fact, this is often the ground truth the model aims to predict.
Instead, we pair each multichannel audio segment with the readily available head pose information, \eg, IMUs.
Since the wearer's pose determines the microphone array's orientation, the model can directly consider self-motion instead of capturing subtle signals about self-motion in audio during training.
These pairs can further be utilized to explicitly synthesize the spatial audio locked to a specific direction if necessary~\cite{pulkki2006directional,ahonen2008directional}.

% \vspace{0.05in}
\noindent
\textbf{Spherical Projection.}
Egocentric field-of-view videos can be visually projected onto a sphere.
This is analogous to a partially observable 360$^\circ$ panorama where the observable region changes with respect to self-motion.
An equiangular mapping 
% $f_{EX}: \mathcal{V}\mapsto\mathcal{V}'$ 
from the $ij$-th pixel in $\mathcal{V}$ to the $XY$-th pixel in $\mathcal{V}'$ can be computed as follows, given horizontal and vertical angular fields-of-view of $\theta_\text{HF}$ and $\theta_\text{VF}$, % a visual input $\mathcal{V} \in \mathbb{R}^{3\times H \times W}$ with 
where  $q \in \mathbb{R}^4$ is the current head pose provided in quaternions, $g_{ij} \in \mathbb{R}^4$ is the spherical world-locked position of $ij$-th pixel in pure quaternions, and $R$ is the radius of a world-locked sphere such that $(H_P,W_p)=(\pi R,2\pi R)$:
\begin{align}
    \label{eq:explicitswl1}
    g_{ij}' &= (0, \tan(\theta_\text{HF} \times(j/W_v-0.5)), \tan(\theta_\text{VF} \times(i/H_v-0.5)), 1), \\
    \label{eq:explicitswl2}
    g_{ij}  &= (0, g_{ij}^x, g_{ij}^y, g_{ij}^z) = q(g_{ij}'/\norm{g_{ij}'}_2)q^{-1}, \\
    \label{eq:explicitwarp}
    (X, Y) &= (R \times \text{atan2}(g_{ij}^z, \sqrt{(g_{ij}^x)^2+(g_{ij}^y)^2}), R \times (\text{atan2}(g_{ij}^y, g_{ij}^x) + \pi)).
\end{align}

% As illustrated in Fig.~\ref{fig:swl}-(a),
% explicit spherical world-locking itself could act as an intuitive near real-time visualization tool for egocentric videos in the wild by canceling out self-motion with the wearer's pose information.
% Since our primary objective at this step is to verify the spatio-temporal consistency of the multisensory streams, it is beyond the scope of this work to further improve precision by introducing additional factors like optical calibration or translation.

Although complete pose information with rotation and translation could be combined to further improve the spatial consistency in Fig.~\ref{fig:swl}-(a), we only consider rotation in light of three observations:
(i) it is more efficient and straightforward to integrate in our model due to the unit quaternion assumption (Eq.~(\ref{eq:spatialsim})),
(ii) rotation suffices for common seated activities like conversations, and 
(iii) the influence of translation becomes negligible within a certain length of time, where we use less than one second in all experiments.

\subsection{Implicit Spherical World-Locking}
\label{subsec:implicitswl}
% \vspace{-0.05in}

While explicit spherical world-locking can be compelling in terms of interaction and visualization, it is less practical to use this representation for model training.
For example, an irregular array access in Eq.~(\ref{eq:explicitwarp}) incurs nontrivial overhead.
Distorted images as in Fig.~\ref{fig:swl}-(a) also introduce another challenge of distortion-aware methods discussed in \S\ref{sec:relatedworks}.
To circumvent these issues, we suggest an implicit way to construct a spherical world-locked representation of multisensory inputs.
As depicted in Fig.~\ref{fig:swl}-(b), we leave all inputs intact and pair them with coordinates on a world-locked sphere to maintain semantic and position embeddings of multisensory inputs separately. % incorporate self-motion separately.

% \vspace{0.05in}
\noindent
\textbf{Multi-CLS Embeddings.}
Classification tokens are commonly employed for capturing the global context from a set of input embeddings.
Since our goal is to localize signals spatially, we exploit multiple classification tokens $\{c_i\}_{i=1}^{N_c}$ parametrized with a point $p_i=(x_i,y_i,z_i)$ on a world-locked sphere to capture semantic information around $p_i$, where $\mathbf{W}_c \in \mathbb{R}^{d\times 3}, \mathbf{b}_c \in \mathbb{R}^d$ are learnable weights:
\begin{equation}
\label{eq:multicls}
    c_i = \mathbf{W}_cp_i+\mathbf{b}_c.
\end{equation}

% Note that increasing complexity in CLS token parametrization, \eg, MLP, did not improve the performance of the model.

Some recent works use multiple CLS tokens for capturing class-specific information in semantic segmentation~\cite{xu2022multi} or ensembling in language understanding~\cite{chang2023multi}.
% ^Unlike previous works, 
The key difference in our work is that 
we deploy multiple CLS tokens to predict spatially localized signals on a sphere for flexible decoding (\S\ref{subsec:decoder}).
% In our framework, multiple classification tokens are deployed to predict localized signals on a sphere, facilitating flexibility in output decoding (\S\ref{subsec:decoder}).

% \vspace{0.05in}
\noindent
\textbf{Semantic Input Embeddings.}
Since implicit spherical world-locking naturally separates position embeddings on a world-locked sphere from semantic embeddings, we can use off-the-shelf feature encoders for processing unmodified multisensory inputs.
% Thanks to implicit spherical world-locking, all inputs are projected to the embedding space using conventional feature encoders without the need for explicit spherical projection.
To obtain visual embeddings $\{v_i\}_{i=1}^{N_v}$, we use ResNet-18~\cite{he2016deep} or 3-layer ConvNet to process frames or facial images.
For audio embeddings $\{a_i\}_{i=1}^{N_a}$, we apply linear projection per patch as in \cite{dosovitskiy2021image,gong2021ast,yun2023dense}, where we use vertical patches of the spectrogram, as shown in Fig.~\ref{fig:swl}-(b), for the audio semantic embeddings to accurately align with the wearer's head pose.
Finally, similar to Eq.~(\ref{eq:multicls}), we assign a learnable embedding for behavioral input $\{b_i\}_{i=1}^{N_b}$, which is also parametrized with a point on a unit sphere.

% \vspace{0.05in}
\noindent
\textbf{Position Input Embeddings.}
% For each semantic input, we assign a point on a world-locked sphere to model spatial dynamics separately, as illustrated in Fig.~\ref{fig:swl}-(b).
For each semantic input embedding, we assign a 3D point on a sphere so that all of the multisensory input embeddings are implicitly located on a world-locked sphere, %to model spatial dynamics separately, 
as illustrated in Fig.~\ref{fig:swl}-(b).
For classification tokens and behavioral inputs, we use the coordinates identical to the ones used in semantic inputs.
For audio and visual inputs, we assign the head orientation $q$ in Eq.~(\ref{eq:explicitswl2}) and spherical world-locked location of each visual token, respectively.
In short, our multisensory input embeddings are summarized as
\begin{align}
\label{eq:input_semantic}
    \{x_i\}_{i=1}^N&=\{x_i^0\}_{i=1}^N=\{c_1,...,c_{N_c}, a_1,...,a_{N_a},v_1,...,v_{N_v},b_1,...,b_{N_b}\}, \\
\label{eq:input_position}
    \{p_i\}_{i=1}^N&=\{(x_1,y_1,z_1),...,(x_N,y_N,z_N)\}.
\end{align}

\section{Multisensory Spherical World-Locked Transformer}
\label{sec:approach_must}

We propose the Multisensory Spherical World-Locked Transformer (MuST) to perform audio-visual localization tasks in egocentric videos, building upon the concept of implicit spherical world-locking and multimodal transformers.
Using multisensory input embeddings in Eq.~(\ref{eq:input_semantic}--\ref{eq:input_position}),
we exploit MuST encoder blocks that focus on multisensory interactions on a world-locked sphere (\S\ref{subsec:encoder}), followed by a lightweight decoder for tackling various localization tasks flexibly (\S\ref{subsec:decoder}).

\begin{figure}[t]
    % \vspace{-5pt}
    \centering
\includegraphics[trim=0.0cm 0.0cm 0cm 0.0cm,clip,width=1\textwidth]{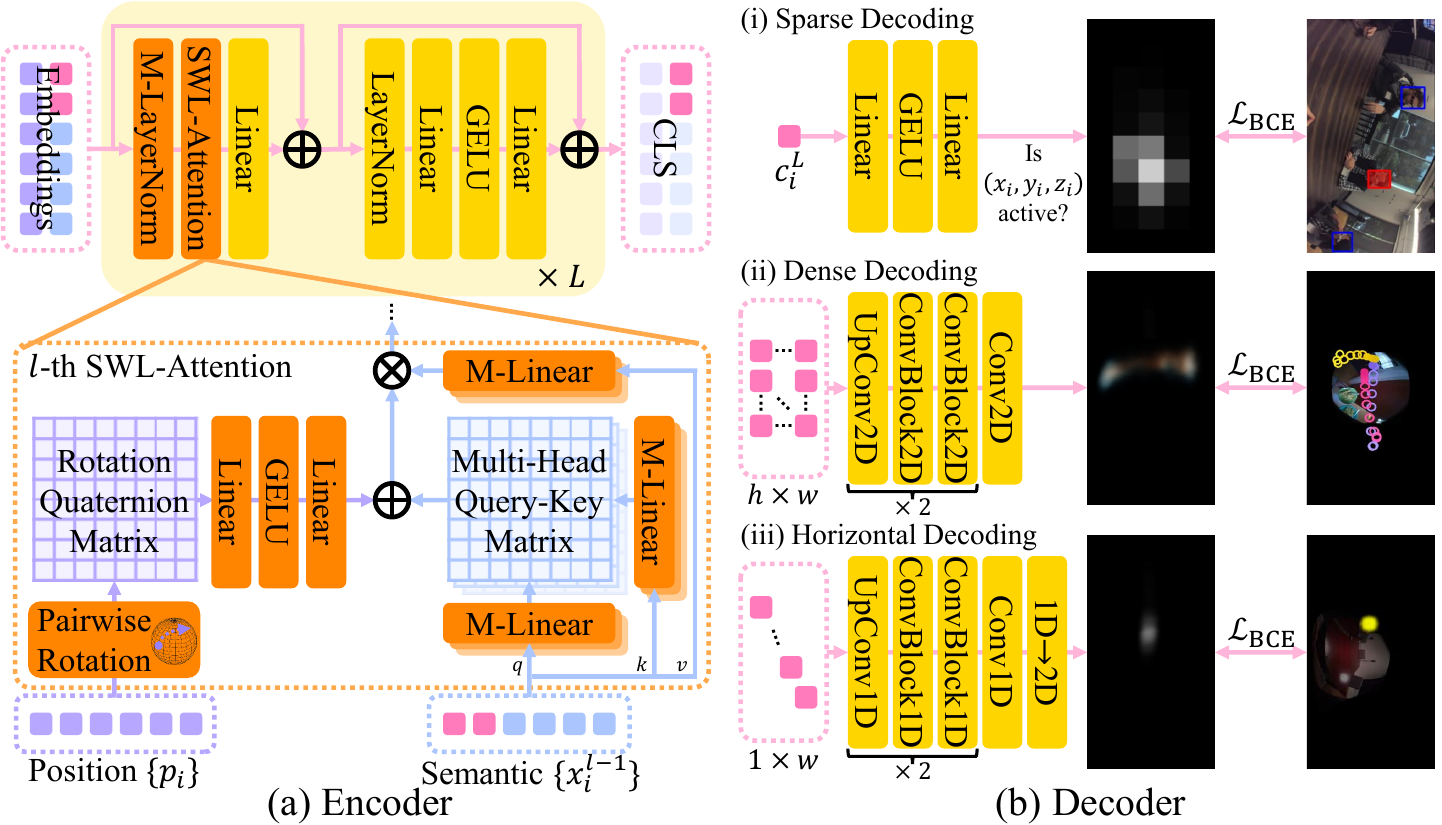}
    \caption{Our MuST model architecture. M- indicates modality-wise operations. 
    }
    \label{fig:architecture}
    % \vspace{-10pt}
\end{figure}
% \RGnote{This figure could be more polished. The yellow arrow on the left for semantic and blue one for spatial on the right are somewhat hard to notice, and I am wondering whether there is a way to reorg the figure a bit. Also, is it necessary to have four separate globe shapes for spatial? Might be a bit redundant.}
% \RGnote{Much better I think. Not sure whether it's a good idea, but another way to make it clearer and less crowded is only showing encoder and decoder as abstract boxes, and make use of the space here to also illustrate the *tasks* conceptually, which currently are not shown in any figures. Then you have a small and separate encoder and decoder figure to show the technical details as Fig 4, or even just put it in Supp. I don't have a strong opinion here.}}

\subsection{MuST Encoder}
\label{subsec:encoder}

% MuST comprises a transformer encoder variant for multisensory integration and a lightweight decoder for localization.
Transformers~\cite{vaswani2017attention} can effectively integrate multisensory embeddings 
and their corresponding positions, forming an ideal combination of position and semantic embeddings from implicit spherical world-locking.
Our MuST block is built upon the general Transformer block \cite{dosovitskiy2021image},
with two key differences in self-attention to incorporate multiple senses on a world-locked sphere: spatial similarity and modality-wise operations.
%, where the gist of spherical world-locking is incorporated in two core aspects

% \vspace{0.05in}
\noindent
\textbf{Spatial Similarity on Sphere.}
Since each multisensory embedding retains a position on a world-locked sphere, we can model the pairwise interaction between different modalities in the form of rotation quaternions.
This spatial similarity matrix at $l$-th layer $\mathbf{P}^l \in \mathbb{R}^{N\times N}$ is integrated into each (semantic) query-key matrix in the multi-head self-attention, promoting spatial relations among embeddings.
The rotation from a 3D unit vector $p_i$ to $p_j$ is computed as
\begin{equation}
\label{eq:spatialsim}
    \mathbf{P}_{ij}^l = \text{Linear}(\text{GELU}(\text{Linear}([1+p_i \cdot p_j,\, p_i\times p_j]))),
\end{equation}
where $(1+p_i \cdot p_j, p_i\times p_j)$ is a rotation quaternion and the output of MLP is scalar, \ie, $\mathbb{R}^4 \rightarrow \mathbb{R}$.
% As depicted in the rotation quaternion matrix from Fig.~\ref{fig:architecture}-(a), learned spatial similarity can capture complex relations between points on a sphere.
Since the pairwise rotation remains identical for all layers, rotation quaternions are computed once for each input and used for all layers.

% \vspace{0.05in}
\noindent
\textbf{Modality-wise Operations.}
Since our goal is to encode a heterogeneous set of modalities in a single encoder, it is paramount to harmonize them during training.
We promote cross-modal interactions in each encoder block by applying layer normalization~\cite{ba2016layer} and $q,k,v$ projection in multi-head attention in a modality-specific manner, \ie, M-LN and M-Attn in Table~\ref{tab:easycom}-(b).
By normalizing the embeddings with modality-wise means and variances while retaining the modality-specific mapping before dot-product attention in each layer, \ie using different linear projection per modality, the model notably promotes an interplay among different modalities than its unimodal counterparts.
Note that not all modality-specific modules positively influence the model training, which is further discussed in \S\ref{subsec:exp_easycom}. %Table~\ref{tab:easycom}-(b).
In short, multi-head self-attention for each head is
\begin{align}
\label{eq:attention}
    \bar{x}_i^{l} &= \text{Linear}(0.5 \times (\sigma(Q^lK^{lT}/\sqrt{d}) + \sigma(\mathbf{P}^l))V^l), \\
    x_i^{l+1} &= x_i^l + \bar{x}_i^{l} + \text{Linear}(\text{GELU}(\text{Linear}(\bar{x}_i^{l}))),
\end{align}
where $\sigma$ denotes softmax and $Q^l,K^l,V^l$ are queries, keys, and values projected with modality-wise linear layers (\ie, M-Linear in Fig.~\ref{fig:architecture}), respectively.
% LayerNorms in Fig.~\ref{fig:architecture}-(b) are omitted for brevity.

\begin{comment}
\begin{figure}[t]
    \centering
\includegraphics[trim=0.0cm 0.0cm 0cm 0.0cm,clip,width=0.95\textwidth]{figures/fig_architecture_decoder.pdf}
    \caption{Target-specific decoding methods of MuST.}
    \label{fig:decoder}
\end{figure}
\end{comment}

\subsection{MuST Decoder}
\label{subsec:decoder}

Using multiple CLS tokens obtained from the last encoder layer, \ie, $\{c_i^L\}_{i=1}^{N_c}$, we can employ different decoding strategies depending on the target task. % the characteristics of the target task

% \vspace{0.05in}
\noindent
\textbf{Sparse Decoding.}
For each token $c_i^L$, we apply pointwise decoding with an MLP to obtain score $y_i$ that corresponds to our model's prediction on point $p_i$:
\begin{equation}
\label{eq:sparsedecoder}
    y_i = \text{Linear}(\text{GELU}(\text{Linear}(c_i^L))).
\end{equation}
Unless mentioned otherwise, we use a sparse grid of $5\times 10$ CLS tokens for training the model, \ie, each token covers around 2\% of the output region.
It is possible to make the training more efficient by selecting a subset of classification tokens for the model like a set of pre-detected regions of interest or faces (\eg, MuST with Sparse Point in Table~\ref{tab:easycom}-(d)).
This reduces the number of CLS tokens by an order of magnitude smaller, which is far more efficient than the full grid.

% \vspace{0.05in}
\noindent
\textbf{Dense Decoding.}
From a grid of CLS tokens, we can utilize a light deconvolutional network illustrated in Fig.~\ref{fig:architecture}-(b)-(ii) to obtain a dense output map $\mathbf{y}$ of the desired resolution in either a field of view ($H_v\times W_v$) or a spherical panorama ($H_p\times W_p$).
One special case of dense decoding is horizontal decoding (Fig.~\ref{fig:architecture}-(b)-(iii)), which is applicable to spherical source localization.
Since the spherical world-locking compensates self-motion in the data stream, our scene representation is gravity-aligned.
So, for some tasks, it is possible to discard tokens except for the ones around the equator line without notable performance degradation.
For example, there is only a marginal performance gap in our model reported in Table~\ref{tab:chat}, \ie, around $1^\circ$, in spite of reducing the number of CLS tokens by a factor of five.
In this case, 1D operations instead of 2D can be used for decoding and converted to 2D by permuting channel dimension to vertical dimension, making both encoding and decoding more efficient.

% \vspace{0.05in}
\noindent
\textbf{Learning Objective and Details.}
We adapt the binary cross entropy loss for training the model, \ie, $\frac{1}{N_c}\sum\mathcal{H}(y_i, \hat{y}_i)$ for sparse decoding and $\frac{1}{H\times W}\sum\mathcal{H}(\mathbf{y}, \mathbf{\hat{y}})$ for dense decoding where $\mathcal{H}$ denotes cross entropy.
We use the Adam optimizer~\cite{kingma2015adam} with a learning rate of 1e-4 without scheduling.
The model is trained end-to-end for 10 epochs until convergence, where we closely follow the hyperparameters used in a smaller variant of Vision Transformer (DeIT-S~\cite{touvron2021training}), which has slightly fewer parameters than ResNet-50~\cite{he2016deep}.

% \vspace{-0.15in}
\section{Experiments}
% \vspace{-0.1in}
\label{sec:experiments}

For a comprehensive demonstration of the effectiveness of our framework, we evaluate MuST with multiple benchmarks covering diverse egocentric videos.
We first focus on egocentric audio-visual active speaker localization (\S\ref{subsec:exp_easycom}). 
% which aims to promptly locate active speakers within the egocentric frame.
In addition, we report the performance of auditory spherical source localization (\S\ref{subsec:exp_chat}) to evaluate the model's capability of localizing directional signals on a sphere without visual shortcuts.
Finally, we generalize our framework to more diverse everyday activities by developing a new suite of tasks on egocentric behavior anticipation (\S\ref{subsec:exp_ariapilot}), which jointly predicts the direction of wearer's future behaviors, \ie, gaze, head orientation, and trajectory, from multisensory inputs.

% \vspace{-0.1in}
\subsection{Audio-Visual Active Speaker Localization}
\label{subsec:exp_easycom}

\textbf{Dataset.}
EasyCom~\cite{donley2021easycom} is
% the first 
a public dataset 
% on multisensory 
of 
egocentric conversations 
for 
% in the 
augmented reality 
% scenario.
applications.
It consists of 0.38M video frames and their corresponding sensory inputs like pose and multichannel audio.
Due to the highly noisy nature of audio streams from the microphone array and frequent self-motion, the dataset covers various challenges in egocentric multi-speaker conversations 
% like beamforming or 
such as
speech enhancement.
We focus on active speaker localization as one of the most common egocentric audio-visual localization tasks. % to validate the model's scene understanding proficiency with multisensory inputs.

\noindent
\textbf{Experiment Settings.}
We closely follow prior works' experiment 
% settings~\cite{jiang2022egocentric,murdock2024self} 
settings~\cite{jiang2022egocentric} 
like splits and metrics for a fair comparison.
We mainly report the mean average precision (mAP), which captures both spatial and temporal localization of speech activity inside the camera's field-of-view.
The mAP scores of all models are computed by pooling the maximum logit value within the corresponding head bounding boxes.
We also devise an Oracle comparison as a potential upper bound on the egocentric models' performance using close microphone recordings from the other participants, which are unavailable from the wearer's perspective, to detect speech activity from cleaner near-field audio.

We compare our full model 
and its variants 
with a number of competitive baselines.
% as well as variants of our model.
We report the performance of a state-of-the-art audio signal processing method~\cite{tourbabin2019direction}, visually oriented methods like mouth region classifier~\cite{jiang2022egocentric} and Dense Prediction Transformer~\cite{ranftl2021vision}, and competitive localization frameworks based on LSTM~\cite{wu2021binaural} or Transformer~\cite{tao2021someone}.
We also report more recent frameworks on egocentric audio-visual 
localization~\cite{jiang2022egocentric,murdock2024self} 
%localization~\cite{jiang2022egocentric} 
for thorough comparison.
Finally, ablation studies on encoder-decoder design, input modalities, and temporal window are provided for a comprehensive analysis of MuST.

\begin{table}[t]
    \caption{
    Performance of active speaker localization on the EasyCom dataset~\cite{donley2021easycom}.
    }
    % \vspace{-10pt}
    \begin{subtable}[t]{0.35\textwidth} \centering
        \begin{tabular}[t]{l|c} \hline
        (a) Methods & mAP$_\uparrow$ \\ \hline
        DOA~\cite{tourbabin2019direction}               & 52.62 \\
        DPT~\cite{ranftl2021vision}                     & 61.66 \\
        MRC~\cite{jiang2022egocentric}                  & 64.24 \\
        BAVNet~\cite{wu2021binaural}                    & 60.75 \\
        TalkNet~\cite{tao2021someone}                   & 69.13 \\
        BPN$_\text{face}$~\cite{murdock2024self}        & 75.22 \\
        AVLN~\cite{murdock2024self}                     & 85.11 \\
        MAVASL$_\text{Spec}$~\cite{jiang2022egocentric} & 85.49 \\
        MAVASL$_\text{C+E}$~\cite{jiang2022egocentric}  & 86.32 \\
        % EgoAVLoc~\cite{huang2023egocentric}             & xx.xx \\ %
        \textbf{MuST}                           & \textbf{89.88} \\ \hline
        Oracle                                          & 91.03 \\ \hline
        \end{tabular}
    \end{subtable}
    \hspace{\fill}
    \begin{subtable}[t]{0.31\textwidth} \centering
        \begin{tabular}[t]{l|c} \hline
        (b) Encoder & mAP$_\uparrow$ \\ \hline
        MuST$_\text{w/o pose}$       & 87.76 \\
        MuST$_\text{w/o rotation}$   & 88.83 \\ \hline
        MuST$_\text{w/o M-ops}$      & 88.53 \\
        MuST$_\text{M-LN}$           & 89.67 \\
        MuST$_\text{M-LN,M-MLP}$     & 89.16 \\
        MuST$_\text{M-LN,M-Attn}$    & \textbf{89.88} \\ \hline
        \multicolumn{2}{c}{} \\ \hline
        (d) Decoder & mAP$_\uparrow$ \\ \hline
        Sparse Point & 88.95 \\
        Dense        & 89.58 \\
        Sparse Grid  & \textbf{89.88} \\ \hline
        \end{tabular}
    \end{subtable}
    \hspace{\fill}
    \begin{subtable}[t]{0.28\textwidth} \centering
        \begin{tabular}[t]{l|c} \hline
        (c) Modality & mAP$_\uparrow$ \\ \hline
        $\mathcal{B}_\text{pose}$                 & 47.95 \\
        +$\mathcal{A}_\text{mono}$                & 68.57 \\
        +$\mathcal{V}$                            & 68.78 \\
        +$\mathcal{A}_\text{mono}$+$\mathcal{V}$  & 73.47 \\
        +$\mathcal{A}_\text{multi}$               & 89.50 \\
        % +$\mathcal{A}$+$\mathcal{V}_\text{frame}$ & 89.16 \\
        +$\mathcal{A}_\text{multi}$+$\mathcal{V}$  & \textbf{89.88} \\ \hline
        \multicolumn{2}{c}{} \\ \hline
        (e) Temporal & mAP$_\uparrow$ \\ \hline
        100ms & 82.96 \\
        200ms & 87.78 \\
        300ms & \textbf{89.88} \\\hline
        \end{tabular}
    \end{subtable}
    \label{tab:easycom}
    % \vspace{-10pt}
\end{table}

% \vspace{0.05in}
\noindent
\textbf{Comparison with Prior Arts.}
In Table~\ref{tab:easycom}-(a), our proposed framework outperforms previous methods by a large margin, increasing the accuracy by 3.6\%p.
This gap becomes wider if the same modalities (faces and spectrograms) are used as inputs, \ie, 4.4\%p. % All models except for BAVNet uses faces as visual input
% A marginal performance gap between AVLN and MuST implies that preserving the original structure of the input stream is preferable to utilizing spherical operations.
%To provide more context regarding the performance gap between our framework and prior arts, we report the mAP of the Oracle for a better grasp of the upper bound in performance.
The performance of our model is comparable with the Oracle's despite the usage of noisy microphone arrays (-1.2\%p).
This tight upper bound is likely due to the coarse-grained nature of annotations in EasyCom, \ie, active speech labels are based on 
phrases instead of phonemes, %phrase-level utterances instead of phoneme-level, 
whereas the model's prediction is evaluated every 50ms.
Such a gap makes it hard for the model to differentiate pauses of active speakers from non-speakers, which is in line with 
performance degradation for
% the degraded performance of models with 
shorter temporal windows in Table~\ref{tab:easycom}-(e).

% \vspace{0.05in}
\noindent
\textbf{Ablation Studies.}
Table~\ref{tab:easycom}-(b) shows the influence of different encoder components on the performance.
Components based on spherical world-locking, like position information and rotation, significantly contribute to the full model's performance, (+2.1\%p).
Also, MuST without modality-specific operations (MuST $_\text{w/o M-ops}$) displays even worse performance than the model without visual information in Table~\ref{tab:easycom}-(c).
Since not all modality-specific operations are beneficial to performance, 
%\eg, the last MLP in MuST block (MuST $_\text{M-LN,M-MLP}$), 
it is essential to reconcile different modalities properly.
%the face-only and mono audio-only models display similar performance, while 
Combining mono audio with visual input introduces a remarkable +4.9 mAP gain, even outperforming a large-scale pretrained prior art for single-channel localization~\cite{tao2021someone}.
Still, the largest performance boost comes from the multichannel microphone array,
which can capture rich spatial signals with multiple microphones.
% It is also worth noting that providing visual signals relevant to the benchmark task, \eg, faces instead of raw frames, leads to a meaningful performance improvement.

\begin{figure}[t]
    \centering
\includegraphics[trim=0.0cm 0.0cm 0cm 0.0cm,clip,width=1\textwidth]{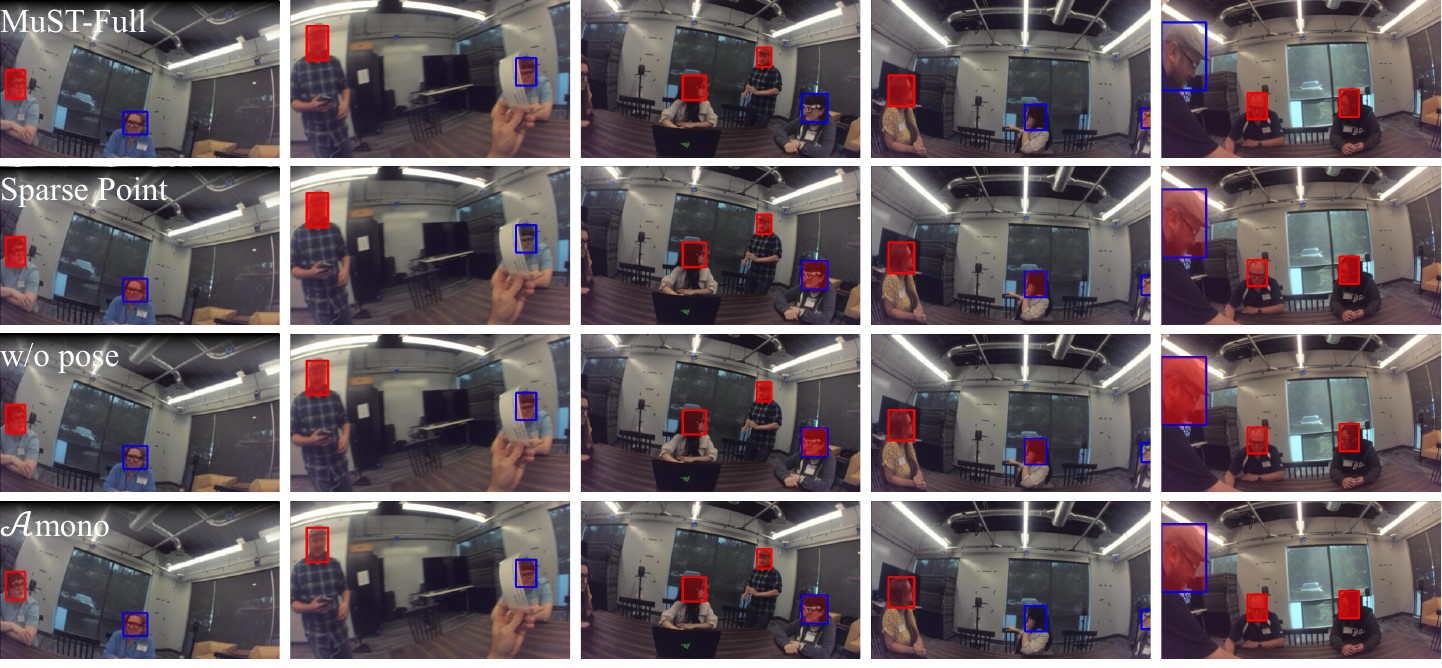}
    % \vspace{-0.2in}
    \caption{
    Qualitative examples of egocentric active speaker localization on EasyCom~\cite{donley2021easycom}. The \textcolor{red}{red}/\textcolor{blue}{blue} boxes indicate active/non-active speakers, and the \textcolor{red}{red} heatmap indicates model prediction.
    MuST can make correct predictions for scenes with gravity misalignment (col. 1), motion blur (col. 2, 4), and multi-speakers (col. 3, 5).
    }
    \label{fig:qual_easycom}
    % \vspace{-10pt}
\end{figure}

% \vspace{0.05in}
\noindent
\textbf{Qualitative Results.}
Fig.~\ref{fig:qual_easycom} compares our full model with selected variants of MuST, demonstrating correct active speaker localization in challenging scenarios with multiple active speakers and diverse self-motion.
The sparse point decoder, which assigns a CLS token to each detected face, is also precise except for the last column with variable head size.
% , where one of the heads appears larger than the others.
Without pose information or multichannel audio, performance falls short due to insufficient spatial reasoning capability.

\subsection{Auditory Spherical Source Localization}
\label{subsec:exp_chat}

\textbf{Dataset.}
%The Reality Labs Research Conversations for Hearing Augmentation Technology (RLR-CHAT) dataset
% RLR-CHAT~\cite{murdock2024self,yin2024hearing} 
RLR-CHAT~\cite{yin2024hearing} 
is a large-scale dataset of egocentric multisensory streams under a variety of configurations, covering an order of magnitude more recordings than in EasyCom~\cite{donley2021easycom}.
RLR-CHAT encompasses more realistic scenarios like scene layouts, overlapping speech in free-form conversations, and varying degrees of background noise.
Due to increased diversity and lower frame rates (\ie, 200ms), it is paramount for the model to perform precise localization as well as disambiguate multiple speakers at the same time.
We particularly focus on a more challenging setup of multichannel auditory source localization that lacks visual cues, which could prevent shortcuts like face regions.

\noindent
\begin{minipage}{\textwidth}
\begin{minipage}[b]{0.49\textwidth}
    \centering
    \captionof{table}{Comparison of auditory spherical source localization errors on the RLR-CHAT 
    % dataset~\cite{murdock2024self,yin2024hearing}.
    dataset~\cite{yin2024hearing}.
    } 
    \begin{tabular}{l|cc} \hline
     & MAE$_{\text{g}\rightarrow\text{p}\downarrow}$ & MAE$_{\text{p}\rightarrow\text{g}\downarrow}$ \\ \hline
    EchoNet~\cite{gao2020visualechoes} & 66.99 & 65.29 \\
    MAVASL-A~\cite{jiang2022egocentric} & 62.25 & 60.70 \\
    SAM~\cite{yun2023dense} & 46.95 & 44.90 \\
    MuST$_\text{w/o pose}$ & 29.33 & 28.55 \\ 
    \textbf{MuST} & \textbf{15.23} & \textbf{12.67} \\ \hline
    \end{tabular}
    \label{tab:chat}
\end{minipage}
\hfill
\begin{minipage}[b]{0.46\textwidth}
    \centering
    \includegraphics[trim=0.0cm 0.0cm 0cm 0.0cm,clip,width=\textwidth,height=0.5\textwidth]{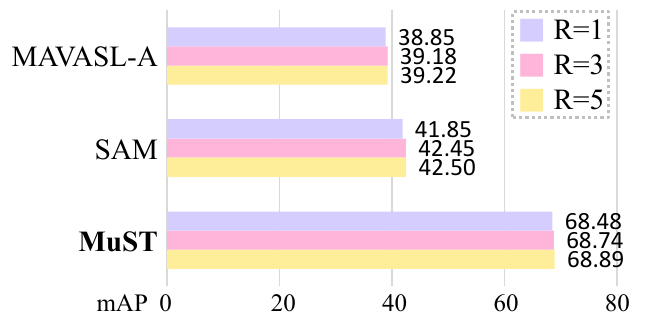}
    \captionof{figure}{Spherical mAP with varying angular precision on 
    % RLR-CHAT~\cite{murdock2024self,yin2024hearing}.
    RLR-CHAT~\cite{yin2024hearing}.
    }
    \label{fig:chat_map}
\end{minipage}
\linebreak % \vspace{10pt}
\end{minipage}

\noindent
\textbf{Experiment Settings.}
Following prior works \cite{jiang2022egocentric}, we report the mean angular error (MAE) from prediction to ground truth and vice versa, reflecting how far the model's prediction deviates from the ground truth source direction on a sphere.
In order to consider the differences among models' output distributions, we select a fixed number of peaks in predictions with non-max suppression, \ie, the number of active speakers in a 200ms timeframe.

We report the performance of several state-of-the-art models on spatial reasoning with audio using identical audio features (multichannel STFT) and ground truth for a fair comparison.
We use the audio network of MAVASL~\cite{jiang2022egocentric}, EchoNet \cite{gao2020visualechoes} for spatial reasoning with echolocation, and the SAM audio network~\cite{yun2023dense} for dense indoor prediction with sound.
We also provide spherical mAP with varying angular resolutions to get a better grasp of spatial precision.
Please refer to \cite{lv2024aria} for details regarding the microphone array configuration used in experiments.

% \vspace{0.05in}
\noindent
\textbf{Performance Analysis.}
Table~\ref{tab:chat} summarizes the performance of different auditory spherical source localization methods.
Our framework achieves superior performance in both MAE metrics with or without pose, and the usage of pose information substantially improves the audio-based localization performance.
Fig.~\ref{fig:chat_map} illustrates the mAP scores on a sphere with varying angular precision, where radius values of 1, 3, and 5 correspond to 
% the precision 
detection thresholds
of 2.25$^\circ$, 6.75$^\circ$, and 11.25$^\circ$, respectively.
Despite generally lower performance 
% than in EasyCom 
due to higher complexity, angular precision variations do not have a notable influence on mAP scores, implying the precise localization ability of correct predictions.

\begin{figure}[t]
    \centering
\includegraphics[trim=0.0cm 0.0cm 0cm 0.0cm,clip,width=1\textwidth]{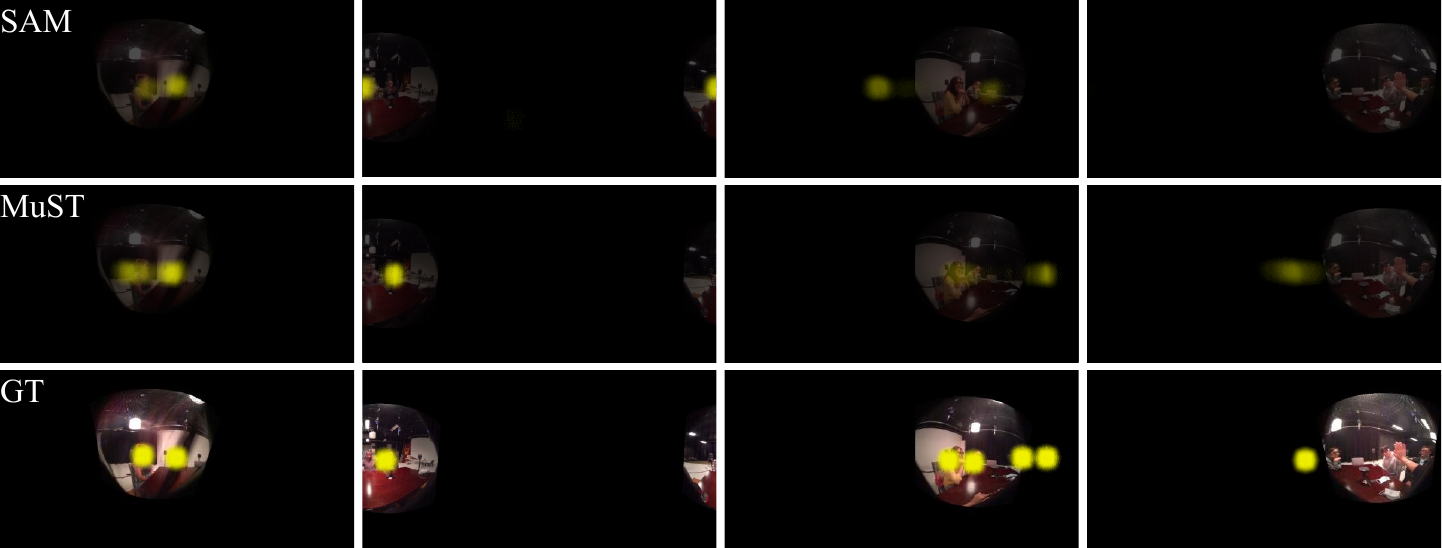}
    \caption{Qualitative examples of auditory spherical source localization on 
    % RLR-CHAT~\cite{murdock2024self,yin2024hearing}. 
    RLR-CHAT~\cite{yin2024hearing}. 
    Our model displays precise detection as well as localization capability over the prior art~\cite{yun2023dense}.
    Note that visual frames are not used in all models.
    }
    \label{fig:qual_chat}
    % \vspace{-5pt}
\end{figure}

Fig.~\ref{fig:qual_chat} depicts qualitative examples of audio-only localization performance, where all models efficiently bypass a visually challenging scenario of a hair occlusion in the first column.
In addition, our framework displays better source detection and localization capability than the competitive baseline~\cite{yun2023dense}.

\begin{table}[t]
    \centering
    \caption{Comparison of behavior anticipation errors on the AEA Dataset~\cite{lv2024aria}.}
    % \vspace{-5pt}
    \begin{tabular}{l|ccc|ccc|ccc} \hline
     & \multicolumn{3}{c|}{Gaze} & \multicolumn{3}{c|}{Orientation} & \multicolumn{3}{c}{Trajectory} \\ \hline
    MAE$_\downarrow$ & 
    $T_\text{300ms}$ & $T_\text{500ms}$ & $T_\text{700ms}$ & 
    $T_\text{300ms}$ & $T_\text{500ms}$  & $T_\text{700ms}$ & 
    $T_\text{300ms}$ & $T_\text{500ms}$ & $T_\text{700ms}$ \\ \hline
    MultitaskGP~\cite{gardner2018gpytorch} & 
    11.42 & 15.59 & 18.40 & \textbf{4.70} &  9.28 & 12.27 & 13.75 & 17.86 & 20.02 \\ \hline
    MuST$_\mathcal{AV}$ &
    12.26 & 14.48 & 16.59 &  5.68 &  8.28 & 10.75 & 92.38 & 92.46 & 92.71 \\
    MuST$_\mathcal{B}$ &
    \textbf{8.78} & 11.98 & 14.65 &  5.02 &  7.65 & 10.18 & \textbf{9.77} & \textbf{12.04} & \textbf{13.48} \\
    MuST$_\mathcal{VB}$ &
     8.92 & \textbf{11.96} & \textbf{14.57} &  4.82 &  7.40 &  9.91 & 10.05 & 12.36 & 13.90 \\
    MuST$_\mathcal{AVB}$ &
     9.17 & 12.15 & 14.75 &  4.78 & \textbf{7.36} & \textbf{9.90} &  9.96 & 12.38 & 13.95 \\
    MuST$_{\mathcal{AVB}\text{-singletask}}$ &
     9.19 & 12.35 & 15.02 &  5.02 &  7.74 & 10.28 & 10.08 & 12.53 & 14.03 \\ \hline
    \end{tabular}
    \label{tab:aria}
    % \vspace{-10pt}
\end{table}

\subsection{Egocentric Behavior Anticipation}
\label{subsec:exp_ariapilot}

\textbf{Dataset.}
We extend MuST to perform multisensory localization in more diverse egocentric daily activities to examine the generalizability of our proposed framework.
The Aria Everyday Activities (AEA) Dataset~\cite{lv2024aria} covers diverse egocentric videos of daily activities like cooking or chatting for scene comprehension, comprising 143 recordings from five different environments.
Understanding and anticipating the wearer's behaviors, \ie, eye gaze, head orientation, and motion trajectory, in daily activities can be crucial in egocentric user understanding and applicable for assistive systems in augmented or mixed reality scenarios.
Since there is no public benchmark that jointly anticipates
a set of cohesive egocentric behaviors with multisensory observations to the best of our knowledge, we organize a suite of tasks for holistic egocentric behavior anticipation with AEA. %the AEA Dataset.

% \vspace{0.05in}
\noindent
\textbf{Experiment Setting.}
We tackle three egocentric behavioral targets on the world-locked sphere: eye gaze, head orientation, and motion trajectory.
% Given current audio-visual observations and previous behavioral contexts of 700ms, our goal is to anticipate future behaviors in 300/500/700ms 
Considering the typical behavioral reaction time of humans,
% ~\cite{murdock2024self}, \ie, 300ms, 
our goal is to anticipate future behaviors in 300/500/700ms given the current audio-visual observations and previous behavioral contexts of 700ms.
% We exclude anticipation at 100ms due to negligible difference with the current context.
To evaluate localization performance, we use Mean Angular Errors (MAE) of behaviors at different timestamps by comparing the argmax coordinate of the model's prediction with ground truth behaviors, similar to \S\ref{subsec:exp_chat}.
We report the average performance from a five-fold cross-validation using five different scenes in the dataset.
As the problem of egocentric behavior anticipation in this dataset has not been addressed previously, we report the performance of a competitive baseline of Multitask Gaussian Process~\cite{gardner2018gpytorch}
%Since the baseline models from previous tasks are either incapable of making such prediction or require considerable modification to use multiple behavioral contexts in 3D vectors as input we test various our model's variants for an extensive analysis of this benchmark.
as well as selected variants of our model for an extensive analysis. %of this benchmark.

% \vspace{0.05in}

\begin{figure}[t]
    \centering
    \includegraphics[trim=0.0cm 0.0cm 0cm 0.0cm,clip,width=1\textwidth]{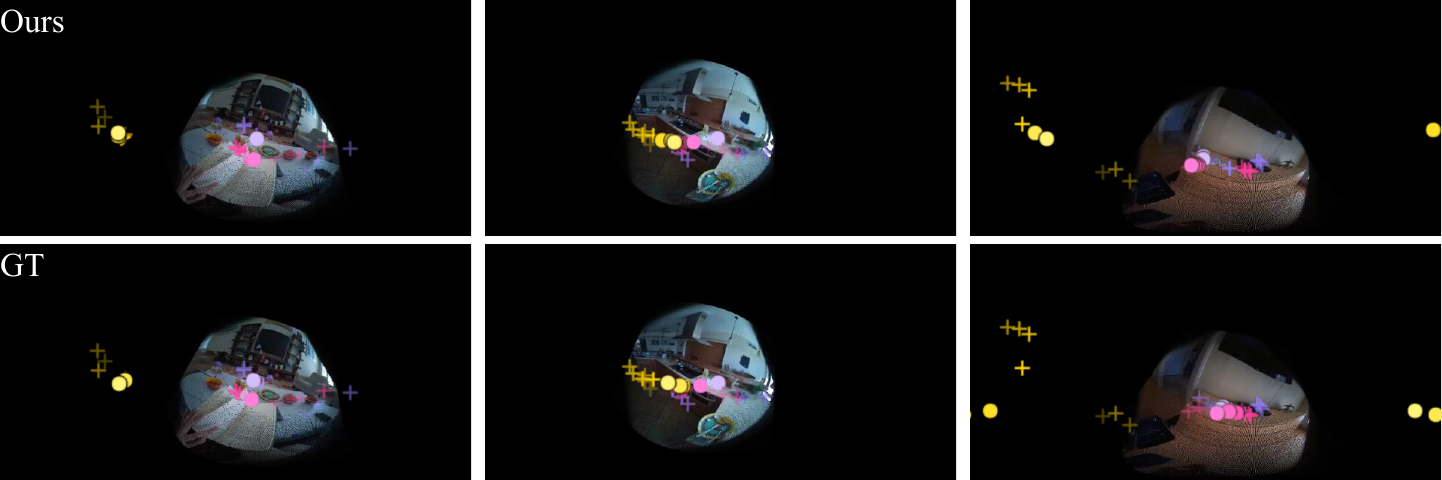}
    % \vspace{-10pt}
    \caption{Qualitative examples of egocentric behavior anticipation on the AEA Dataset~\cite{lv2024aria} where \textcolor{MainPurple}{gaze}, \textcolor{MainPink}{orientation}, and \textcolor{MainYellow}{trajectory} are color coded.
    Cross/circle symbols denote previous/anticipated behaviors.
    Our model reasonably anticipates future behaviors in common scenarios like long-term fixation and human-object interaction.
    }
    \label{fig:qual_aria}
    % \vspace{-10pt}
\end{figure}

\noindent
\textbf{Performance Analysis.}
As visualized in Fig.~\ref{fig:qual_aria}, egocentric behavior is quite complex and often challenging to predict.
Still, our model achieves consistent performance improvements over the baseline except for short-term head orientation anticipation, reducing the angular error by 20.5\%, 12.8\%, and 29.5\% for gaze, orientation, and trajectory respectively.
It is noteworthy that our audio-visual model without previous gaze context (MuST$_\mathcal{AV}$) displays compelling performance in gaze anticipation task but is poor at predicting future head trajectories.
Such tendency suggests that, unlike exocentric behaviors, egocentric behaviors like gaze can be anticipated from current audio-visual observations to a meaningful extent without previous behavioral context.
Different sets of input modalities are often more proficient for a specific task than others.
For example, audio inputs are closely tied with previous pose information, achieving better performance in anticipating orientation than others.
Lastly, our model trained to jointly anticipate all behaviors outperforms single-task counterparts in all tasks,  meaning that MuST is properly leveraging coherence across different behavioral contexts.
% We defer further discussion to Appendix.

\section{Conclusion}
\label{sec:conclusion}

We presented the Spherical World-Locking, a new framework for audio-visual localization in egocentric videos
%addressing the challenges of self-motion in egocentric videos for audio-visual localization.
that leverages the wearer's pose information to offset challenges in self-motion. % and also enables learning better multisensory scene representation.
Powered by implicit SWL, our MuST architecture facilitates cross-modal interaction on a world-locked sphere by means of rotation quaternions and modality-wise operations, enabling learning better multisensory scene representation. It also provides fine-grained and flexible decoding for localization with multiple spatial classification tokens.
We have conducted extensive experiments on three different multisensory egocentric localization benchmarks. Our results demonstrate significant improvement both quantitatively and qualitatively over prior arts. %across diverse metrics. Furthermore, our approach exhibits qualitatively remarkable performance in localizing various source signals from multisensory observations.
As future work, we plan to extend our framework to exploit other modalities like optical flow as a proxy of pose information, which is not always available, and scale to more large-scale egocentric video datasets. 

%to large-scale egocentric video datasets, where the pose information is occasionally unavailable, by exploiting other modalities like optical flow as a proxy.

% \clearpage

\bibliographystyle{splncs04}
\bibliography{eccv2024_swl}
\end{document}